\documentclass[fleqn,10pt]{wlscirep}
\usepackage[utf8]{inputenc}
\usepackage[T1]{fontenc}
\usepackage{subfig}
\usepackage{graphicx}
\usepackage{amsmath}
\usepackage{nccmath}
\usepackage{xspace}
\makeatletter
\DeclareRobustCommand\onedot{\futurelet\@let@token\@onedot}
\def\@onedot{\ifx\@let@token.\else.\null\fi\xspace}
\def\etal{\emph{et al}\onedot}
\makeatother

\usepackage{lineno}
\title{SwinCheX: Multi-label classification on chest X-ray images with transformers
}

\author[1,+]{Sina Taslimi}
\author[1,+,*]{Soroush Taslimi}
\author[1]{Nima Fathi}
\author[1]{Mohammadreza Salehi}
\author[1]{Mohammad Hossein Rohban}
\affil[1]{Sharif University of Technology, Computer Engineering department, Tehran, Iran}
\affil[*]{Correspondence and requests for materials should be addressed to M.H. Rohban (rohban@sharif.edu)}

\affil[+]{these authors contributed equally to this work}

\begin{abstract}
According to the considerable growth in the avail of chest X-ray images in diagnosing various diseases, as well as gathering extensive datasets, having an automated diagnosis procedure using deep neural networks has occupied the minds of experts. Most of the available methods in computer vision use a CNN backbone to acquire high accuracy on the classification problems. Nevertheless, recent researches show that transformers, established as the de facto method in NLP, can also outperform many CNN-based models in vision. This paper proposes a multi-label classification deep model based on the Swin Transformer as the backbone to achieve state-of-the-art diagnosis classification. It leverages Multi-Layer Perceptron, also known as MLP, for the head architecture. We evaluate our model on one of the most widely-used and largest x-ray datasets called “Chest X-ray14”, which comprises more than 100,000 frontal/back-view images from over 30,000 patients with 14 famous chest diseases. Our model has been tested with several number of MLP layers for the head setting, each achieves a competitive AUC score on all classes. Comprehensive experiments on Chest X-ray14 have shown that a 3-layer head attains state-of-the-art performance with an average AUC score of 0.810, compared to the former SOTA average AUC of 0.799. We propose an experimental setup for the fair benchmarking of existing methods, which could be used as a basis for the future studies. Finally, we followed up our results by confirming that the proposed method attends to the pathologically relevant areas of the chest.
\end{abstract}
\begin{document}

\flushbottom
\maketitle
\thispagestyle{empty}

\iffalse
\noindent Please note: Abbreviations should be introduced at the first mention in the main text – no abbreviations lists. Suggested structure of main text (not enforced) is provided below.
\fi

\section*{Introduction}
Disburdening medical experts of monotonous tasks such as medical imaging diagnosis have been a long-standing goal in bioinformatics. Moreover, with the emergence of a global pandemic, there is an unprecedented increasing interest in reliable, accurate, and fast methods for diagnosing Chest X-ray (CXR) images. In this pursuit, Neural Network-based Computer Vision algorithms have proven to be successful at similar tasks, by surpassing human-level performance, both in terms of accuracy and speed (Goodfellow \etal \cite{pmlr-v28-goodfellow13}, Srivastava \etal \cite{brainstorm2015}, Sabour \etal \cite{NIPS2017_2cad8fa4} and Mazzia \etal \cite{mazzia2021efficient}). Training these networks for the CXR diagnosis usually requires large amounts of labeled/annotated data, which is recently provided to the community by the National Institute of Health (NIH). This multi-labeled dataset, “ChestX-ray14”\cite{Wang_2017}, consists of above 100k frontal/back-view X-ray images and spurred us to tackle this problem.

Several attempts have been made to detect abnormalities in ChestX-ray14. Wang \etal\cite{Wang_2017} assessed performance of some of the most well-known Convolutional Neural Network (CNN) architectures (e.g., AlexNet\cite{NIPS2012_c399862d}, VGGNet\cite{simonyan2015deep}, ResNet\cite{he2015deep}) that are pre-trained on ImageNet\cite{deng2009imagenet}. Yao \etal\cite{yao2018weakly} exploited dependencies among disparate labels with an RNN-based model using Long-Short Term Memory Networks (LSTM)\cite{hochreiter1997long} along with a CNN backbone adopting a variant of DenseNet\cite{huang2018densely}, which barely was trained entirely on ChestX-ray14\cite{Wang_2017}. Rajpurkar \etal\cite{rajpurkar2017chexnet} enhanced a DenseNet-121\cite{huang2018densely} with transfer-learning to predict multi-label outputs. Eventually, Guendel \etal\cite{guendel2018learning} proposed a location-aware method called DNetLoc grounded on DenseNet-121\cite{huang2018densely}, where an early stopping procedure was employed to avoid overfitting.

Unfortunately, a trusty and fair comparison among all different methods has been virtually impossible. In many early works, dissimilar dataset splits of training and test sets have been employed. First and foremost, a particular patient could exist in both training and test data which, according to the correlation between different images of the same patient, could yield a wrong and inflated detection rate. As a result, Wang \etal\cite{Wang_2017} released a new official patient-wise split later. Yao \etal\cite{yao2018weakly}, Guendel \etal\cite{guendel2018learning}, and Baltruschat \etal\cite{Baltruschat2019} published their evaluations based on this official split. For instance, to highlight the effect of the training split on evaluations, Guendel \etal\cite{guendel2018learning} state that average AUC score of their model falls roughly 3 percent when using the official data split.

% One of the most significant subjects in clinical contexts is that every medical community and patients have to trust machines' predictions. The presented method must provide visual evidence to supports the results.

Figure \ref{fig:arch} provides an overview of our approach. 
In this paper, we focus on multi-label classification using Vision Transformers (ViT)\cite{dosovitskiy2020vit}. We adapt a pre-trained Swin Transfomer\cite{liu2021swin} model and attach multi-head feedforward neural network layers for each class. We assess our model on four distinct configurations (e.g., with no head, one head, two heads, and three heads). Furthermore, to validate the results further, we obtained Grad-CAM saliency heatmaps for few input images to confirm that the model pays attention to pathologically relevant areas of the chest.

%as a novel work, we represent attention (AT) for the Swin Transformer\cite{liu2021swin} model to illustrate which part of the image is responsible for the decision. 
According to evaluations on the aforenamed dataset (i.e., Chest X-ray14\cite{Wang_2017}), quantitative results have shown that the proposed model achieves state-of-the-art result. In other words, we attain the best average AUC score, as well class-wise AUC on various classes, based on the reports on the Chest X-Ray14 dataset\cite{Wang_2017}.

\section*{Background}
For transparency, in this section, we discuss some deep learning concepts. We start with RNNs and then discuss their existing drawbacks. We continue with transformers and how they solved these issues, and how we can use transformers in vision-based tasks such as ours.
\subsection*{Recurrent Neural Networks}
RNNs are a class of neural networks that allow previous outputs to be used as inputs while summarizing all given information in the hidden states. They are broadly used to model sequential dependencies. There are various RNN methods introduced for solving the problem of memorizing long-term dependencies, such as Long short-term memory networks, also known as LSTM \cite{hochreiter1997long}, and gated recurrent units (GRU) \cite{chung2014empirical}. Although RNNs are among the most substantial state-of-the-art architectures for sequence modeling, there are some issues regarding these networks:
\begin{itemize}
    \item Sequential processing of the input prohibits parallelization within instances.
    \item There is a famous problem in RNNs around vanishing and exploding gradients, which is an essential field of study in the research community.
    \item Long-range patterns are still poorly modeled.
    \item Distance between time positions are implicitly assumed to be linear.
    \item There is no explicit modeling of long and short-range dependencies.
\end{itemize}

\subsection*{Attention}
The attention mechanism is one of the most valuable breakthroughs in the last decade, and has spawned the rise of so many recent advances in deep models such as transformers \cite{NIPS2017_3f5ee243}. The attention mechanism attempts to selectively concentrate on a few relevant aspects of the input while ignoring others in deep neural networks. The idea is that not only all the inputs should participate in forming the context vector, but also their relative importance should be taken into account. Therefore, whenever the model generates an output, it searches the whole hidden states in the encoder to find the most pertinent information. \cite{bahdanau2016neural}.

\subsection*{Self-Attention}
Cheng \etal \cite{cheng2016long} defined self-attention as bonding different positions of a single sequence to attain a more explicit representation. A key vector of dimension $d_{k}$, a query vector of dimension $d_{k}$, and a value vector of dimension $d_{v}$, which are an abstraction of the embedding vectors in different subspaces, are defined for each embedding of the words. To compute the attention of a target word concerning the input embeddings, we multiply the query of the target and the key of the input word. Then, these matching scores act as the weights of the value vectors during the summation.
\\
In practice, we compute the attention function on a set of queries simultaneously, with three packed matrices Q, K, and V for queries, keys, and values as follows:
\begin{ceqn}
\begin{align}
Attention(Q, K, V) = Softmax \left({\frac {Q K^{T}}{\sqrt{d_{k}}}}\right) V
\label{eq:1}
\end{align}
\end{ceqn}
To wrap the merits of the attention:
\begin{itemize}
    \item There is a constant path length between any two positions.
    \item Parallelization is trivial (per layer).
    \item It can replace the aligned recurrence entirely.
\end{itemize}

\subsection*{Transformers}
Transformers are self-attention-based architectures that are trying to solve the problem of parallelization by using convolutional neural networks in tandem with attention models\cite{NIPS2017_3f5ee243}. In general, attention boosts how fast the model can translate from one sequence to another. The architecture of these networks is mainly composed of one encoder and one decoder section with a linear layer on top of them. The encoder comprises a stack of six layers, each consisting of two sub-layers. The first sub-layer is a multi-head self-attention, and the second is a basic, fully connected feed-forward layer. The decoder section is similar to the encoder with six analogous layers; each layer contains three sub-layers. The first is much like the encoder, a multi-head attention layer. Following these layers, we have a multi-head attention layer that performs attention over the output of the encoder stack. At last, there is a fully-connected feed-forward layer just like the encoder stack.
\\
To be noticed, there is a sub-layer called multi-head attention in transformers, which is slightly different from the self-attention. The intuition behind it, which is inspired by filters in convolutional neural networks, is that one may pay extra attention to some specific words in translating a word based on the type of question being asked. In multi-head attention, a unique linear transformation with a different query, key, and value vector for each word embedding is represented for each head to learn different relationships. 

\subsection*{Vision Transformer}
 Mesmerized by Transformers, several previous work combined CNNs with self-attention (Wang \etal \cite{Wang_nonlocalCVPR2018}, Carion \etal \cite{carion2020endtoend}). Others just replaced the whole convolution layers with the attention mechanism (Ramachandran \etal \cite{NEURIPS2019_3416a75f},Wang \etal, 2020a \cite{wang2020axial}). Most of the transformer-based models could not be successfully scaled on modern hardware until Dosovitskiy \etal came up with the idea of Vision Transformer (ViT)\cite{dosovitskiy2020vit}. Vision Transformer uses the original architecture of Transformer in the work of Vaswani \etal \cite{NIPS2017_3f5ee243} for machine translation with just dropping the decoder part. Transformers lack inductive bias in the form of translation equivariance. In fact,  they are permutation invariant by design and cannot process grid-structure data. Dosovitskiy \etal \cite{dosovitskiy2020vit} converted the non-sequential spatial images into sequences by defining patches.
 \\
 The standard transformer model in Vaswani \etal \cite{NIPS2017_3f5ee243} processes a 1D sequence of token embeddings as input. To use this notion, for an input image $ x \in R^{H \times W \times C} $ and patch size $ p \times p $, we create N patches with $ N = \frac{H \times W}{P \times P} $ (same as the sequence length in sentences). Each patch  $ x_{p} \in R^{P \times P \times C} $ needs to be transformed into a 1D vector $ d \in R^D $ with a linear transformation layer to form embeddings. The images are then transformed into 1D embedding vectors through the self-attention. To this end, we also add positional embeddings, to account for the patches relative order. To obtain the output, we feed the sequence as an input to the standard transformer encoder, and then pass the output to the final MLP layers for the classification. 
\\
In practice, ViTs are pre-trained on huge fully-supervised datasets, and are fine-tuned to the downstream tasks. To successfully fine-tune the pre-trained model, essentially, the original prediction MLP layer is replaced by a zero-initialized D$\times$K feed-forward layer (with k = number of classes in the downstream task). It has been shown that ViT would beat or at least approach state-of-the-art CNN models in various vision tasks if pre-trained on massive datasets like ImageNet-21k and JFT-300M.      

\subsection*{Swin Transformer}
ViT showed promising performance, but there are still some challenges regarding transferring the exceptional performance of transformers in the language domain to the computer vision. These challenges can be demonstrated by the existing distinction between these two modalities. Unlike the tokens that work as essential language processing components, different visual elements can vary significantly in scale, a crucial problem in some vision tasks such as object detection. In previous transformer-based models, we have seen that the tokens are all of a fixed size, a property that is problematic for these vision tasks. Another critical difference between language and vision is the high resolution of pixels in images. Numerous computer vision tasks such as semantic segmentation need a pixel-level dense prediction. The computational complexity in existing transformer models is quadratic in the image size, and hence it is impossible to use them in such architectural forms. Swin transformers were proposed to deal with the mentioned challenges, and operate as a general-purpose backbone for various vision tasks \cite{liu2021swin}.
\begin{figure*}
  \centering
  \includegraphics[width=\linewidth, scale=0.1]{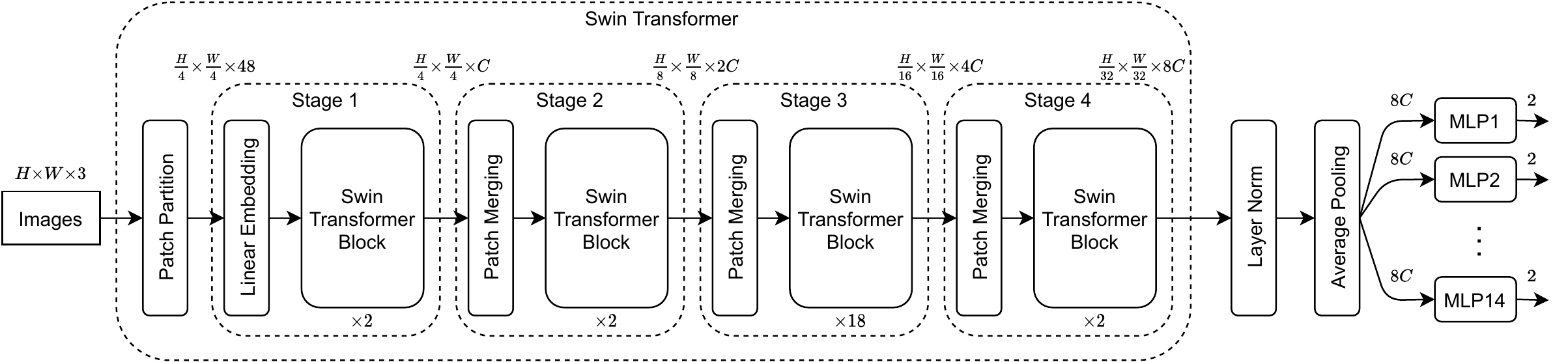}
  \caption{The overall architecture of our method. Chest X-ray images after being converted to RGB and resized to 224 are passed through a Swin-L\cite{liu2021swin} transformer. This follows a Layer-Norm, and a $7\times 7$ average pooling, before the shared section is finished. 14 MLP heads are branched from the shared section. Each MLP head contains 3 layers of 384, 48, and 48 neurons, respectively.}
  \label{fig:arch}
\end{figure*}
\section*{Methods}

% Topical subheadings are allowed. Authors must ensure that their Methods section includes adequate experimental and characterization data necessary for others in the field to reproduce their work.
\subsection*{Dataset}
In this study, we use the ChestX-ray14 dataset \cite{Wang_2017}. This dataset contains 112,120 frontal/back-view chest X-ray images with the size of 1024$\times$1024 (8-bits gray-scale values) from 30,850 patients, with labels from 14 pathology classes. Each image can have multiple labels. The labels are collected by analyzing radiology reports, and are expected to have over 90\% accuracy. For a more accurate and objective comparison, we apply the official patient-wise split gathered by Wang \etal \cite{Wang_2017}

\subsection*{Proposed Method}
% Transformers have recently become a popular method in computer vision.
An overview of our model is provided in the Figure \ref{fig:arch}. We use the Swin transformer as the component that is shared between the models that are predicting each label. To elaborate, the Swin transformer,  like ViT, first splits the input RGB image into non-overlapping patches. Each resulting patch is defined as a token. We use a patch size of $4 \times 4$, and thus the feature dimension is $4 \times 4 \times 3 = 48$. Then a linear embedding layer is applied to this raw-valued feature to project it to an arbitrary dimension of C. After that, we apply several Swin transformer blocks. These blocks maintain the number of tokens ($\frac{H}{4} \times \frac{W}{4}$). As the network gets deeper in layers, the number of tokens decreases by patch merging layers after each stack of the Swin transformer blocks. The first patch merging layer concatenates the feature of each group of $2 \times 2$ neighboring patches and applies a linear layer on the connected features, and the output dimensions are set to 2C. Swin transformer blocks are applied afterward for feature transformation, with the resolution kept at $ \frac{H}{8} \times \frac{W}{8}$. This procedure is repeated two more times, resulting in resolutions of $ \frac{H}{16} \times \frac{W}{16}$ and $\frac{H}{32} \times \frac{W}{32}$, respectively. 
\subsection*{Swin Transformer Block}

A Swin transformer block is built by replacing the standard multi-head self-attention (MSA) module with a module based on shifted windows, keeping other layers the same. The standard transformer architecture leverages global self-attention, where the associations between a token and all other tokens are calculated. This procedure tends to have a quadratic complexity with respect to the number of tokens, which is immensely infeasible for many vision tasks that involve high-resolution images. To solve this issue, self-attention is computed {\it within} local windows. The windows divide the image in a non-overlapping way. Assuming each window has $M \times M$ patches, the computational complexity of a global multi-head self-attention module and a window-based one of an image of $h \times w$ patches are: 
\begin{ceqn}
\begin{align}
\Omega(MSA) = 4hwC^2 + 2 \left( hw \right)^2C 
\label{globalBased}
\end{align}
\end{ceqn}

\begin{ceqn}
\begin{align}
\Omega(W\--MSA) = 4hwC^2 + 2M^2  hwC
\label{windowBased}
\end{align}
\end{ceqn}
The global MSA has a time complexity that is quadratic to the number of patches, making it inapplicable for a high-resolution image. In contrast, the window-based technique runs in linear time in terms of the number of patches, when M is fixed. 

However, the lack of connection between windows in the window-based self-attention technique limits its power and flexibility. Shifted window partitioning method was suggested to additionally  model the cross window connections. The first module leverages a regular window partitioning scheme, which starts from the top-left pixel. Then, the next module uses a windowing setup, shifted  $M/2 \times M/2$ pixels from that of the regularly partitioned windows of the proceeding layer. Consecutive Swin transformer blocks are computed as:
\begin{ceqn}
\begin{align}
\hat{z}^l = W\--MSA\left(LN(z^{l-1})\right) + z^{l-1}, 
\label{z1}
\end{align}
\end{ceqn}
\begin{ceqn}
\begin{align}
z^l = MLP\left(LN(\hat{z}^l)\right) + \hat{z}^l,
\label{z2}
\end{align}
\end{ceqn}
\begin{ceqn}
\begin{align}
\hat{z}^{l+1} = SW\--MSA\left(LN(z^l)\right) + z^l, \label{z3}
\end{align}
\end{ceqn}
\begin{ceqn}
\begin{align}
z^{l+1} = MLP\left(LN(\hat{z}^{l+1}\right) + \hat{z}^{l+1},
\label{z4}
\end{align}
\end{ceqn}
where $\hat{z}^{l}$ and $z^{l}$ denote the output features of the (S)W-MSA modules, and the MLP module for the block l, respectively. Furthermore, LN denotes the layer normalization module.

\subsection*{Multi-task Learning}

Motivated by the multi-prediction head architecture for the multi-task learning problems, a shared architecture is placed at the front and one head is considered for each pathology after the front section.
We use Swin-L \cite{liu2021swin} transformer as the shared section. The model weights are initialized from ImageNet22k\cite{deng2009imagenet} pre-trained weights.
Swin-L outputs 1536 channels of $7\times7$ feature vectors. To reduce the output to a 1-dimensional vector and utilize consequent fully-connected layers, a Layer-Norm and a $7\times7$ average pooling is applied to the output.
The architecture used in the head sections is the Multi-Layer Perceptron. For our main model, the 3-layer headed SwinCheX, we use a 3-Layer MLP with 384, 48, 48 neurons in each layer, respectively. The last layer is connected to one output, which is normalized by a sigmoid function. Therefore, each head outputs a number between zero and one, showing the probability of that sample containing the corresponding disease. For training of the network, binary cross-entropy loss is considered as the loss function. 32 is used as the batch size in training and the learning rate is set to $3\times10^{-5}$.
Figure \ref{fig:arch} shows the overall architecture of our proposed method.

\subsection*{Evaluation} \label{sec:evaluation}
Some of the previous works in this field used some hyper-parameters without providing any reasons for the choice of the specified values. For example, Guendel \etal\cite{guendel2018learning} used a learning decay of 10x when the validation loss plateaued. Seyyed-Kalantari \etal\cite{seyyedkalantari2020chexclusion} decreased the learning rate by a factor of 2 if the validation loss did not improve over 3 epochs, and they stopped the model if it did not improve over 10 epochs. Furthermore, they fine-tuned the degree of random rotation data augmentation from a set of pre-defined values. While the general idea behind these approaches seems reasonable, the problem is that these hyper-parameters are fine-tuned to perform well on the validation set and therefore, there is the possibility that these hyper-parameters overfit the validation set.
Even more so, these unprincipled hyper-parameter setups may have also been optimized to perform well on the given test split. Therefore, by changing the dataset, slightly modified hyper-parameters result in better performance. Furthermore, code reproducibility is one of the important factors in practical problems. With more hyper-parameters and more conditions on early stopping, reproduction of the results becomes more challenging.

To prevent the mentioned problems, we define a general evaluation method. In our proposed evaluation protocol that is followed in all of our experiments, we split the official training data into training and validation sets, with 80 percent of the set forming the training split. No patient appears in both training and validation splits, for the purpose of prevention of biased prediction of the models. The area under the curve of the receiver operating characteristic (AUROC) is considered as the evaluation metric of the model. The model is trained on the aforementioned training split, and AUC is reported for the validation set after each epoch. Then, the network from the epoch with the maximum validation AUC is selected and its AUC on the official test set is reported. Therefore, in this approach, few hyper-parameters are used and no unclear values for early stopping hyper-parameters are utilized. In this matter, our model selection method is much simpler and can easily be applied to any other dataset. Our train validation split, as well as our code, are available in our Github repository (\href{https://github.com/rohban-lab/SwinCheX}{https://github.com/rohban-lab/SwinCheX}), which allows for effortless reproduction of results.

\section*{Experiments}
Several experiments are run on different transformers to evaluate transformers' performance on the Chest-xRay14 dataset. Moreover, we test the previous state-of-the-art work in the field on our data split to make a fair comparison.

% another subsection?
In all of our experiments, we observe that AUC boosts up in the early epochs, but starts to decline beyond a certain point. This observation holds in all of our experiments, including the already known models in the field such as DNet \cite{guendel2018learning} and ResNet \cite{Baltruschat2019}. This happens because these models consist of millions of parameters and after a point, they begin to overfit to the training set. Furthermore, a good starting point for the optimization process is important to speed up the process and find a better local minimum. In the following, we introduce these experiments and discuss the results.

\subsection*{ViT (Vision Transformer)}
In this experiment, the ViT vision transformer\cite{dosovitskiy2020vit} is tested on each pathology present in the ChestX-ray14 dataset. This experiment is used to give an insight on how well transformers can operate on chest X-ray images, and a comparison is made with our proposed method and other works in table \ref{table:Models_AUC}.

A single neuron is considered as the vision transformer output, and a sigmoid activation function is applied so as to represent the probability of that pathology existence. Weights are initialized from the network that is pre-trained on ImageNet22k\cite{deng2009imagenet}. The weights are then fine-tuned during training on CXR images. Batch size and learning rate are set to 32 and $3\times10^{-2}$, respectively.

\subsection*{Swin Transformer}
Swin transformer constitutes the shared section of our proposed model. Therefore, multiple scenarios are tested with the Swin transformer.

First, a plain Swin transformer with no head attachments is tested. We may also refer to this model as the headless SwinCheX model in the following sections. In the headless case, 14 neurons are considered as the Swin transformer output, and the binary cross-entropy loss is considered as the loss function. In this approach, the whole network is shared for all of the pathologies, and therefore the model flexibility for learning specific functions for each pathology becomes significantly limited.

In order for the model to learn different, and more complex functions for each pathology, an MLP head structure is attached to the output of the Swin transformer. The head structures have no shared parameters in common and therefore, enable the model to learn more complex functions that are specific to each pathology. In this case, the Swin transformer remains as the shared structure between all pathologies. This approach enables the model to learn low-level features of images, which are common between all pathologies, through the shared section, and learn  high-level features that are specific to each pathology in the head sections.

Consequently, in the next set of experiments, the impact of different head structures on the prediction accuracy is investigated. We examine various structures with a different number of layers in the MLP head section. The general structure of the model in these experiments is similar to the main method, except for the MLP heads. In particular, removing MLP layers from our main method is experimented on, so as to check the effectiveness of having more layers. Comparisons between head structure containing one layer of 48 neurons, a head structure containing two layers of 384 and 48 neurons respectively, and the proposed method that contains three layers of 384, 48, and 48 neurons respectively, are provided in Table \ref{table:Models_AUC}, alongside the aforementioned headless architecture. Hyper-parameters, such as learning rate and batch size, are set the same in all the experiments, in order to make a fair comparison.

\subsection*{DenseNet}
DenseNets are widely used in the chest X-ray classification tasks. For instance, they are used in the work by Guendel \etal\cite{guendel2018learning} and Seyyed-Kalantari \etal\cite{seyyedkalantari2020chexclusion}. It was observed that in these methods, a significant number of hyper-parameters are present, which results in difficulty in reproduction and generalization. To have a fair comparison, we apply the DenseNet-121 variant model, which was introduced by Guendel \etal\cite{guendel2018learning}, simply known as DNet, on our data split and evaluate this method with the mentioned evaluation approach in the Methods section. Other than the evaluation approach, settings are the same as the ones used in Guendel \etal\cite{guendel2018learning}

\subsection*{Localization}
We provide Grad-CAM \cite{selvaraju2017grad} heatmaps to assess what parts of the lungs our model pays the most attention to diagnose the disease. This mapping can help the expert to spot  potentially important areas for each disease in the input image. To get the Grad-CAM images for the Swin Transformer, we calculate the gradient of the most dominant logit with respect to the activation map that is located after the first layer norm of the last Swin Transformer block.
% add more details?

\begin{table}[t]
\centering
\caption{Comparison of AUC on the ChestX-Ray14 official test set. Last five columns show our tested methods that utilize vision transformers.}
% \vspace{0.1cm}
\scalebox{0.86}{
\begin{tabular}{|c|c|c|c|c|c|c|c|c|c|}
\hline
pathology       & Wang \etal\cite{Wang_2017} & Yao \etal\cite{yao2018weakly} & DNet\cite{guendel2018learning}   & ResNet-38\cite{Baltruschat2019} & ViT\cite{dosovitskiy2020vit}            & headless & 1-layer head     & 2-layer head   & 3-layer head   \\
       &  &  &            &   &          & SwinCheX & SwinCheX     & SwinCheX   & SwinCheX   \\
\hline
\hline
Cardiomegaly    & 0.81        & 0.856      & 0.883          & 0.875          & \textbf{0.891} & 0.871    & 0.867          & 0.865          & 0.875          \\
Emphysema       & 0.833       & 0.842      & 0.895          & 0.895          & 0.832          & 0.905    & 0.889          & 0.906          & \textbf{0.914} \\
Edema           & 0.805       & 0.806      & 0.835          & 0.846          & 0.842          & 0.844    & 0.848          & \textbf{0.851} & 0.848          \\
Hernia          & 0.872       & 0.775      & 0.896          & \textbf{0.937} & 0.867          & 0.829    & 0.834          & 0.803          & 0.855          \\
Pneumothorax    & 0.799       & 0.805      & 0.846          & 0.84           & 0.838          & 0.869    & 0.867          & 0.864          & \textbf{0.871} \\
Effusion        & 0.759       & 0.806      & \textbf{0.828} & 0.822          & 0.813          & 0.826    & 0.827          & 0.823          & 0.824          \\
Mass            & 0.693       & 0.777      & 0.821          & 0.82           & 0.788          & 0.815    & 0.823          & \textbf{0.834} & 0.822          \\
Fibrosis        & 0.786       & 0.743      & 0.818          & 0.816          & 0.787          & 0.818    & \textbf{0.828} & 0.823          & 0.826          \\
Atelectasis     & 0.7         & 0.733      & 0.767          & 0.763          & 0.743          & 0.777    & 0.774          & 0.775          & \textbf{0.781} \\
Consolidation   & 0.703       & 0.711      & 0.745          & 0.749          & 0.695          & 0.734    & \textbf{0.754} & 0.752          & 0.748          \\
Pleural Thicken & 0.684       & 0.724      & 0.761          & 0.763          & 0.755          & 0.769    & 0.778          & \textbf{0.782} & 0.778          \\
Nodule          & 0.669       & 0.724      & 0.758          & 0.747          & 0.703          & 0.771    & 0.761          & \textbf{0.783} & 0.78           \\
Pneumonia       & 0.658       & 0.684      & \textbf{0.731} & 0.714          & 0.665          & 0.722    & 0.717          & \textbf{0.731} & 0.713          \\
Infiltration    & 0.661       & 0.673      & 0.709          & 0.694          & 0.687          & 0.706    & 0.702          & \textbf{0.711} & 0.701          \\
\hline
Mean            & 0.745       & 0.761      & 0.807          & 0.806          & 0.779          & 0.804    & 0.805          & 0.807          & \textbf{0.81} \\
\hline
\end{tabular}
}
\label{table:Models_AUC}
\end{table}

\iffalse
\begin{figure*}
  \centering
  \includegraphics[width=\linewidth, scale=0.1]{./resources/val-test-AUCs/swin_nih_multihead-1_head.pdf}
  \caption{The overall architecture of our method. Chest X-ray images after being converted to RGB and resized to 224 are passed through a Swin-L\cite{liu2021swin} transformer. Then follows a Layer-Norm and a $7\times 7$ average pooling at the end of the shared section. 14 MLP heads are branched from the shared section. each MLP head contains 3 layers of 384, 48, 48 neurons respectively.}
  \label{fig:swin1_head}
\end{figure*}
\fi

\begin{figure*}
    \centering
        \subfloat[SwinCheX headless]{
            \label{fig:Swin_AUC_headless}
            \includegraphics[width=0.5\textwidth]{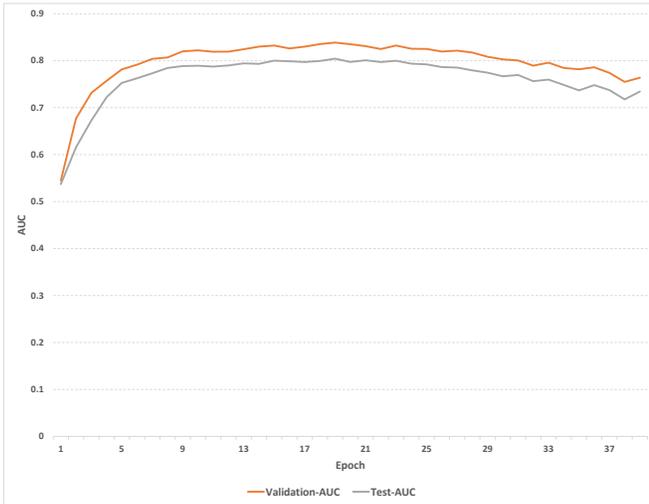}
        }
        \subfloat[SwinChex with a one-layer head]{
            \label{fig:Swin_AUC_1head}
            \includegraphics[width=0.5\textwidth]{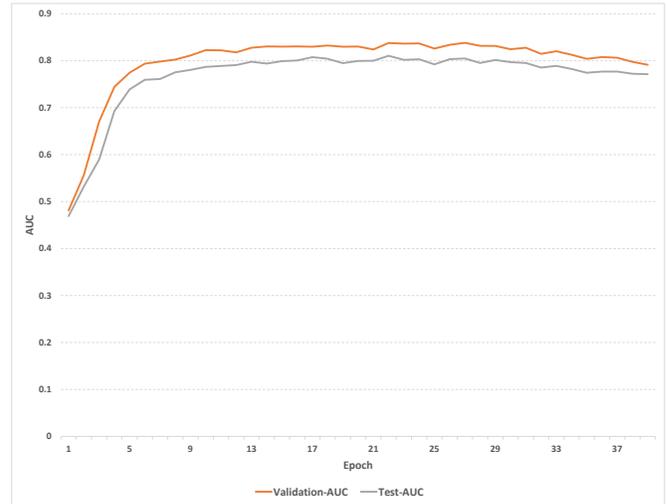}
        }
        \hfill
        \subfloat[SwinCheX with a 2-layers head]{
            \label{fig:Swin_AUC_2head}
            \includegraphics[width=0.5\textwidth]{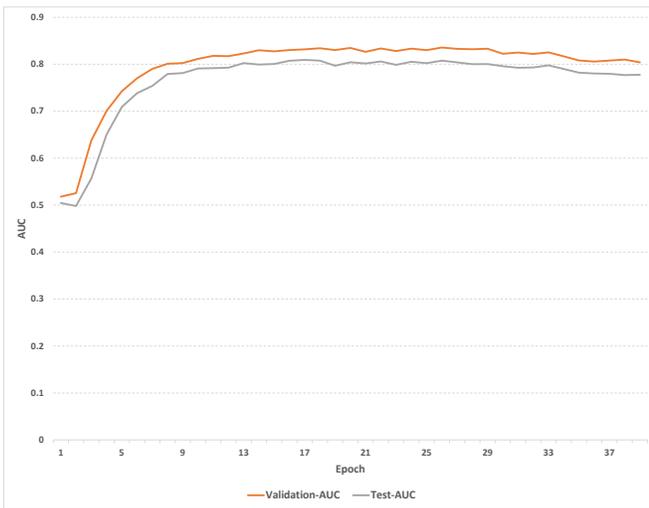}
        }
        \subfloat[SwinCheX with a 3-layers head]{
            \label{fig:Swin_AUC_3head}
            \includegraphics[width=0.5\textwidth]{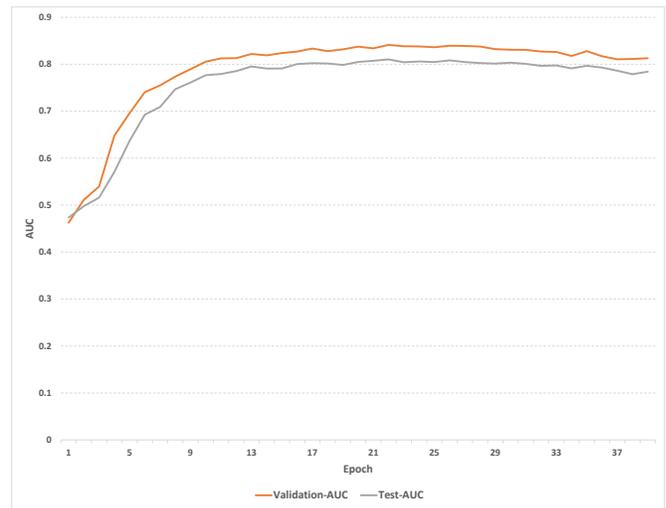}
        }
        \caption{AUC along the training in our proposed model on the validation and test sets. The charts are drawn for 38 epochs. The AUC slightly declines after a certain point in training, which necessitates a model selection method. For this purpose, in all of the settings, the model weights from the epoch that produces the highest AUC in the validation set are chosen.}
        \label{fig:Swin_AUC}
\end{figure*}

\section*{Results}
Detection AUC of existing models in the field, as well as our proposed models are presented in Table \ref{table:Models_AUC}. The maximum value for each pathology, among all the models, is bolded. It is observed that our 3-layer headed SwinCheX model achieves  state-of-the-art AUC value of 0.81 in 14 pathologies on average. Furthermore, SwinCheX with three different head settings achieves  state-of-the-art AUC in 11 pathologies, namely Emphysema, Edema, Pneumothorax, Mass, Fibrosis, Atelectasis, Consolidation, Pleural Thickening, Nodule, Pneumonia, and Infiltration, two of which are achieved by the 1-layer headed SwinCheX, six by the 2-layer headed SwinCheX and three by the 3-layer headed SwinCheX. Also, the ViT transformer reaches a 0.891 AUC score in Cardiomegaly, which is the highest score for this pathology. These results show the power of  multi-resolution transformers in  detecting lung diseases from the X-ray images.

Figure \ref{fig:Swin_AUC} shows the AUC of SwinCheX with different head settings during the training. The orange line shows AUC reached in our validation split and the gray line shows the AUC if the model is evaluated on the test split. These charts are only drawn for demonstration purposes and we do not use any information related to the test split in our model selection. As stated in the evaluation section, we use the epoch with the highest validation AUC to select our model. In these charts, we can see that there is a high correlation between the validation AUC and the test AUC. 
%This result indicates that hyper-parameter fine-tuning is not involved in the selection process because by having the hyper-parameters fine-tuned on the validation set, the model is selected to have the best outcome on the validation set, and therefore, it will not show the same results when tested on a new and unseen dataset. 
Consequently, our approach for selecting the model with the highest validation AUC, would make the selected model perform well in unseen test sets.

\begin{figure*}
    \centering
        \subfloat[Consolidation]{
            \label{fig:GradCAM_Consolidation}
            \includegraphics[width=0.5\textwidth]{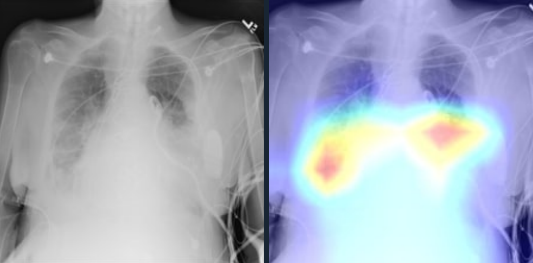}
        }
        \subfloat[Atelectasis]{
            \label{fig:GradCAM_Atelectasis}
            \includegraphics[width=0.5\textwidth]{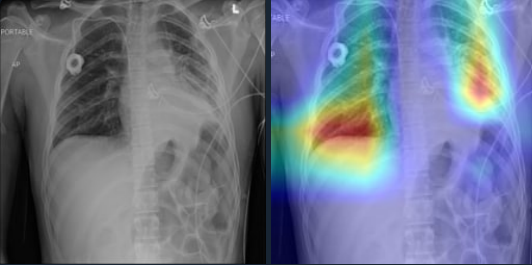}
        }
        \hfill
        \subfloat[Mass]{
            \label{fig:GradCAM_Mass}
            \includegraphics[width=0.5\textwidth]{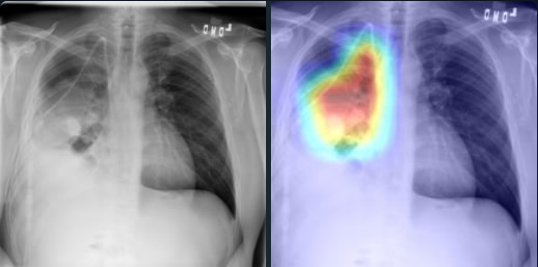}
        }
        \subfloat[Cardiomegaly]{
            \label{fig:GradCAM_Cardiomegaly}
            \includegraphics[width=0.5\textwidth]{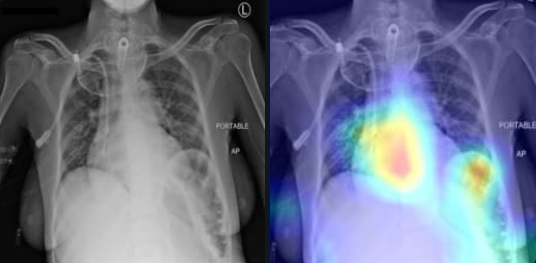}
        }
        \caption{Grad-CAM of our model for positive samples containing Consolidation, Atelectasis, Mass, or Cardiomegaly. For each sample, the original image is shown on the left and the Grad-CAM of the 3-layer headed SwinCheX is shown on the right.}
        \label{fig:GradCAM}
\end{figure*}

% TODO results of localization (grad-cam)
The Grad-CAM mappings of our 3-layer headed SwinCheX for samples containing Consolidation, Atelectasis, Mass or Cardiomegaly are provided in the Figure \ref{fig:GradCAM}. Lung consolidation can be easily seen on an X-ray image by the naked eye. The consolidated parts of the lung look white, or opaque, on a chest X-ray. In Figure \ref{fig:GradCAM_Consolidation} our model detects these parts and diagnoses the sample as positive. Furthermore, an X-ray image can be helpful in the diagnosis of Atelectasis. Findings on an X-ray suggestive of Atelectasis include displacement of fissures, rib crowding, the elevation of the ipsilateral diaphragm, volume loss on ipsilateral hemithorax, hilar displacement, and compensatory hyperlucency of the remaining lobes. In Figure \ref{fig:GradCAM_Atelectasis}, our model has diagnosed this disease by looking at the bottom of the two lungs and their relative positions.

In Figure \ref{fig:GradCAM_Mass} the proposed model diagnoses the sample as a positive case for lung mass by detecting the right area. A lung mass is defined as an abnormal spot or area in the lungs larger than 3 centimeters (cm), about 1.5 inches, in size. Spots smaller than 3 cm in diameter are considered lung nodules. Finally, in Figure \ref{fig:GradCAM_Cardiomegaly}, we have provided a positive case of Cardiomegaly, which is also known as an enlarged heart. In this case, the Grad-CAM mapping of our model shows that it was able to detect this condition by determining where the heart is located and especially, some edge parts of the heart.

\subsection*{Comparison with Previous State-of-the-Art}
The DNet model proposed by Guendel \etal\cite{guendel2018learning} reaches the 0.807 AUC, which presents the previous state-of-the-art in the task of multi-label classification on chest X-ray images. Our 3-layer headed SwinCheX model surpasses this by achieving an AUC score of 0.81. Furthermore, DNet only has a better performance in Cardiomegaly, Hernia, and Effusion than all the different variations of SwinCheX, and our models are dominant in the rest of the diseases. Another major advantage that our model has over DNet is that DNet uses an adaptive learning rate while our model does not utilize this method. Using adaptive learning rates add some hyper-parameters to the model, such as when and by how much one should decrease the learning rate. Finding a good value for these hyper-parameters not only needs a substantial amount of time to train and experiment, but it can also result in overfitting to the validation set, that is, values are selected so that the model performs better on the validation set. Our model does not have these problems and is simpler. To better compare the results of these two models, we tested a DNet with our presented evaluation protocol. The results are shown in the Table \ref{table:SOTA_Comparison_AUC}. We have also put the 3-layer headed SwinCheX results as a reference. It can be seen that DNet with our proposed evaluation method performs worse than DNet on average over all of the pathologies, and our 3-layer headed SwinCheX outperforms it by 1.01\% margin on the AUC.
% maybe also write the results and not just the diff?

\begin{table}[ht]
\centering
\caption{Comparison of the DNet and proposed method based on the AUCs}
\begin{tabular}{|c|c|c|c|}
\hline
pathology       & DNet\cite{guendel2018learning}           & DNet with the proposed & 3-layer head \\
       &            & evaluation method & SwinCheX \\ \hline
Cardiomegaly    & \textbf{0.883} & 0.878                                    & 0.875                 \\
Emphysema       & 0.895          & 0.886                                    & \textbf{0.914}        \\
Edema           & 0.835          & 0.828                                    & \textbf{0.848}        \\
Hernia          & 0.896          & \textbf{0.9}                             & 0.855                 \\
Pneumothorax    & 0.846          & 0.842                                    & \textbf{0.871}        \\
Effusion        & \textbf{0.828} & 0.826                                    & 0.824                 \\
Mass            & 0.821          & 0.803                                    & \textbf{0.822}        \\
Fibrosis        & 0.818          & 0.823                                    & \textbf{0.826}        \\
Atelectasis     & 0.767          & 0.763                                    & \textbf{0.781}        \\
Consolidation   & 0.745          & 0.744                                    & \textbf{0.748}        \\
Pleural Thicken & 0.761          & 0.768                                    & \textbf{0.778}        \\
Nodule          & 0.758          & 0.742                                    & \textbf{0.78}         \\
Pneumonia       & \textbf{0.731} & 0.7                                      & 0.713                 \\
Infiltration    & \textbf{0.709} & 0.686                                    & 0.701                 \\ \hline
Mean            & 0.807          & 0.799                                    & \textbf{0.81}         \\ \hline
\end{tabular}
\label{table:SOTA_Comparison_AUC}
\end{table}

\iffalse
\subsection*{Subsection}

Example text under a subsection. Bulleted lists may be used where appropriate, e.g.

\begin{itemize}
\item First item
\item Second item
\end{itemize}

\subsubsection*{Third-level section}
 
Topical subheadings are allowed.
\fi

\section*{Discussion}

In this paper, we proposed several models based on the vision transformers to perform multi-label classification on the ChestX-ray14 dataset, which contains 14 pathologies of the chest. We also compare the results of these methods and the previous models. We introduce an evaluation method, which we use to select and evaluate the model based on, and for a fair comparison, test the previous state-of-the-art by this evaluation method.

We have shown the efficiency of vision transformers for the task of multi-label classification on the chest X-ray images. We particularly tested and experimented with ViT\cite{dosovitskiy2020vit} and Swin\cite{liu2021swin} transformers. Our results suggest superiority of multi-resolution ViTs in a multi-task learning setup for this task. We propose multi-head MLPs to model the multi-task learning nature of this problem. For future works, other vision transformers can be used and tested. Moreover, the ChestX-ray14 dataset contains images with scattered lung positions, watermarks, and different contrasts. working on a pre-processing technique to remove these issues in an efficient manner before the training process.

% The Discussion should be succinct and must not contain subheadings.

\bibliography{main}

\begin{thebibliography}{10}
\urlstyle{rm}
\expandafter\ifx\csname url\endcsname\relax
  \def\url#1{\texttt{#1}}\fi
\expandafter\ifx\csname urlprefix\endcsname\relax\def\urlprefix{URL }\fi
\expandafter\ifx\csname doiprefix\endcsname\relax\def\doiprefix{DOI: }\fi
\providecommand{\bibinfo}[2]{#2}
\providecommand{\eprint}[2][]{\url{#2}}

\bibitem{pmlr-v28-goodfellow13}
\bibinfo{author}{Goodfellow, I.}, \bibinfo{author}{Warde-Farley, D.},
  \bibinfo{author}{Mirza, M.}, \bibinfo{author}{Courville, A.} \&
  \bibinfo{author}{Bengio, Y.}
\newblock \bibinfo{title}{Maxout networks}.
\newblock In \bibinfo{editor}{Dasgupta, S.} \& \bibinfo{editor}{McAllester, D.}
  (eds.) \emph{\bibinfo{booktitle}{Proceedings of the 30th International
  Conference on Machine Learning}}, vol.~\bibinfo{volume}{28} of
  \emph{\bibinfo{series}{Proceedings of Machine Learning Research}},
  \bibinfo{pages}{1319--1327} (\bibinfo{publisher}{PMLR},
  \bibinfo{address}{Atlanta, Georgia, USA}, \bibinfo{year}{2013}).

\bibitem{brainstorm2015}
\bibinfo{author}{Greff, K.}, \bibinfo{author}{Srivastava, R.~K.} \&
  \bibinfo{author}{Schmidhuber, J.}
\newblock \bibinfo{title}{{Brainstorm: Fast, Flexible and Fun Neural Networks,
  Version 0.5}} (\bibinfo{year}{2015}).

\bibitem{NIPS2017_2cad8fa4}
\bibinfo{author}{Sabour, S.}, \bibinfo{author}{Frosst, N.} \&
  \bibinfo{author}{Hinton, G.~E.}
\newblock \bibinfo{title}{Dynamic routing between capsules}.
\newblock In \bibinfo{editor}{Guyon, I.} \emph{et~al.} (eds.)
  \emph{\bibinfo{booktitle}{Advances in Neural Information Processing
  Systems}}, vol.~\bibinfo{volume}{30} (\bibinfo{publisher}{Curran Associates,
  Inc.}, \bibinfo{year}{2017}).

\bibitem{mazzia2021efficient}
\bibinfo{author}{Mazzia, V.}, \bibinfo{author}{Salvetti, F.} \&
  \bibinfo{author}{Chiaberge, M.}
\newblock \bibinfo{journal}{\bibinfo{title}{Efficient-capsnet: capsule network
  with self-attention routing}}.
\newblock {\emph{\JournalTitle{Scientific reports}}}
  \textbf{\bibinfo{volume}{11}} (\bibinfo{year}{2021}).

\bibitem{Wang_2017}
\bibinfo{author}{Wang, X.} \emph{et~al.}
\newblock \bibinfo{journal}{\bibinfo{title}{Chestx-ray8: Hospital-scale chest
  x-ray database and benchmarks on weakly-supervised classification and
  localization of common thorax diseases}}.
\newblock {\emph{\JournalTitle{2017 IEEE Conference on Computer Vision and
  Pattern Recognition (CVPR)}}} \doiprefix\url{10.1109/cvpr.2017.369}
  (\bibinfo{year}{2017}).

\bibitem{NIPS2012_c399862d}
\bibinfo{author}{Krizhevsky, A.}, \bibinfo{author}{Sutskever, I.} \&
  \bibinfo{author}{Hinton, G.~E.}
\newblock \bibinfo{title}{Imagenet classification with deep convolutional
  neural networks}.
\newblock In \bibinfo{editor}{Pereira, F.}, \bibinfo{editor}{Burges, C. J.~C.},
  \bibinfo{editor}{Bottou, L.} \& \bibinfo{editor}{Weinberger, K.~Q.} (eds.)
  \emph{\bibinfo{booktitle}{Advances in Neural Information Processing
  Systems}}, vol.~\bibinfo{volume}{25} (\bibinfo{publisher}{Curran Associates,
  Inc.}, \bibinfo{year}{2012}).

\bibitem{simonyan2015deep}
\bibinfo{author}{Simonyan, K.} \& \bibinfo{author}{Zisserman, A.}
\newblock \bibinfo{title}{Very deep convolutional networks for large-scale
  image recognition} (\bibinfo{year}{2015}).
\newblock \eprint{1409.1556}.

\bibitem{he2015deep}
\bibinfo{author}{He, K.}, \bibinfo{author}{Zhang, X.}, \bibinfo{author}{Ren,
  S.} \& \bibinfo{author}{Sun, J.}
\newblock \bibinfo{title}{Deep residual learning for image recognition}
  (\bibinfo{year}{2015}).
\newblock \eprint{1512.03385}.

\bibitem{deng2009imagenet}
\bibinfo{author}{Deng, J.} \emph{et~al.}
\newblock \bibinfo{title}{Imagenet: A large-scale hierarchical image database}.
\newblock In \emph{\bibinfo{booktitle}{2009 IEEE conference on computer vision
  and pattern recognition}}, \bibinfo{pages}{248--255}
  (\bibinfo{organization}{Ieee}, \bibinfo{year}{2009}).

\bibitem{yao2018weakly}
\bibinfo{author}{Yao, L.}, \bibinfo{author}{Prosky, J.},
  \bibinfo{author}{Poblenz, E.}, \bibinfo{author}{Covington, B.} \&
  \bibinfo{author}{Lyman, K.}
\newblock \bibinfo{title}{Weakly supervised medical diagnosis and localization
  from multiple resolutions} (\bibinfo{year}{2018}).
\newblock \eprint{1803.07703}.

\bibitem{hochreiter1997long}
\bibinfo{author}{Hochreiter, S.} \& \bibinfo{author}{Schmidhuber, J.}
\newblock \bibinfo{journal}{\bibinfo{title}{Long short-term memory}}.
\newblock {\emph{\JournalTitle{Neural computation}}}
  \textbf{\bibinfo{volume}{9}}, \bibinfo{pages}{1735--1780}
  (\bibinfo{year}{1997}).

\bibitem{huang2018densely}
\bibinfo{author}{Huang, G.}, \bibinfo{author}{Liu, Z.},
  \bibinfo{author}{van~der Maaten, L.} \& \bibinfo{author}{Weinberger, K.~Q.}
\newblock \bibinfo{title}{Densely connected convolutional networks}
  (\bibinfo{year}{2018}).
\newblock \eprint{1608.06993}.

\bibitem{rajpurkar2017chexnet}
\bibinfo{author}{Rajpurkar, P.} \emph{et~al.}
\newblock \bibinfo{title}{Chexnet: Radiologist-level pneumonia detection on
  chest x-rays with deep learning} (\bibinfo{year}{2017}).
\newblock \eprint{1711.05225}.

\bibitem{guendel2018learning}
\bibinfo{author}{Guendel, S.} \emph{et~al.}
\newblock \bibinfo{title}{Learning to recognize abnormalities in chest x-rays
  with location-aware dense networks} (\bibinfo{year}{2018}).
\newblock \eprint{1803.04565}.

\bibitem{Baltruschat2019}
\bibinfo{author}{Baltruschat, I.~M.}, \bibinfo{author}{Nickisch, H.},
  \bibinfo{author}{Grass, M.}, \bibinfo{author}{Knopp, T.} \&
  \bibinfo{author}{Saalbach, A.}
\newblock \bibinfo{journal}{\bibinfo{title}{Comparison of deep learning
  approaches for multi-label chest x-ray classification}}.
\newblock {\emph{\JournalTitle{Scientific Reports}}}
  \textbf{\bibinfo{volume}{9}}, \bibinfo{pages}{6381},
  \doiprefix\url{10.1038/s41598-019-42294-8} (\bibinfo{year}{2019}).

\bibitem{dosovitskiy2020vit}
\bibinfo{author}{Dosovitskiy, A.} \emph{et~al.}
\newblock \bibinfo{journal}{\bibinfo{title}{An image is worth 16x16 words:
  Transformers for image recognition at scale}}.
\newblock {\emph{\JournalTitle{ICLR}}}  (\bibinfo{year}{2021}).

\bibitem{liu2021swin}
\bibinfo{author}{Liu, Z.} \emph{et~al.}
\newblock \bibinfo{title}{Swin transformer: Hierarchical vision transformer
  using shifted windows} (\bibinfo{year}{2021}).
\newblock \eprint{2103.14030}.

\bibitem{chung2014empirical}
\bibinfo{author}{Chung, J.}, \bibinfo{author}{Gulcehre, C.},
  \bibinfo{author}{Cho, K.} \& \bibinfo{author}{Bengio, Y.}
\newblock \bibinfo{journal}{\bibinfo{title}{Empirical evaluation of gated
  recurrent neural networks on sequence modeling}}.
\newblock {\emph{\JournalTitle{arXiv preprint arXiv:1412.3555}}}
  (\bibinfo{year}{2014}).

\bibitem{NIPS2017_3f5ee243}
\bibinfo{author}{Vaswani, A.} \emph{et~al.}
\newblock \bibinfo{title}{Attention is all you need}.
\newblock In \bibinfo{editor}{Guyon, I.} \emph{et~al.} (eds.)
  \emph{\bibinfo{booktitle}{Advances in Neural Information Processing
  Systems}}, vol.~\bibinfo{volume}{30} (\bibinfo{publisher}{Curran Associates,
  Inc.}, \bibinfo{year}{2017}).

\bibitem{bahdanau2016neural}
\bibinfo{author}{Bahdanau, D.}, \bibinfo{author}{Cho, K.} \&
  \bibinfo{author}{Bengio, Y.}
\newblock \bibinfo{title}{Neural machine translation by jointly learning to
  align and translate} (\bibinfo{year}{2016}).
\newblock \eprint{1409.0473}.

\bibitem{cheng2016long}
\bibinfo{author}{Cheng, J.}, \bibinfo{author}{Dong, L.} \&
  \bibinfo{author}{Lapata, M.}
\newblock \bibinfo{title}{Long short-term memory-networks for machine reading}
  (\bibinfo{year}{2016}).
\newblock \eprint{1601.06733}.

\bibitem{Wang_nonlocalCVPR2018}
\bibinfo{author}{Wang, X.}, \bibinfo{author}{Girshick, R.},
  \bibinfo{author}{Gupta, A.} \& \bibinfo{author}{He, K.}
\newblock \bibinfo{title}{Non-local neural networks}.
\newblock In \emph{\bibinfo{booktitle}{CVPR}} (\bibinfo{year}{2018}).

\bibitem{carion2020endtoend}
\bibinfo{author}{Carion, N.} \emph{et~al.}
\newblock \bibinfo{title}{End-to-end object detection with transformers}
  (\bibinfo{year}{2020}).
\newblock \eprint{2005.12872}.

\bibitem{NEURIPS2019_3416a75f}
\bibinfo{author}{Ramachandran, P.} \emph{et~al.}
\newblock \bibinfo{title}{Stand-alone self-attention in vision models}.
\newblock In \bibinfo{editor}{Wallach, H.} \emph{et~al.} (eds.)
  \emph{\bibinfo{booktitle}{Advances in Neural Information Processing
  Systems}}, vol.~\bibinfo{volume}{32} (\bibinfo{publisher}{Curran Associates,
  Inc.}, \bibinfo{year}{2019}).

\bibitem{wang2020axial}
\bibinfo{author}{Wang, H.} \emph{et~al.}
\newblock \bibinfo{title}{Axial-deeplab: Stand-alone axial-attention for
  panoptic segmentation}.
\newblock In \emph{\bibinfo{booktitle}{European Conference on Computer Vision
  (ECCV)}} (\bibinfo{year}{2020}).

\bibitem{seyyedkalantari2020chexclusion}
\bibinfo{author}{Seyyed-Kalantari, L.}, \bibinfo{author}{Liu, G.},
  \bibinfo{author}{McDermott, M.}, \bibinfo{author}{Chen, I.~Y.} \&
  \bibinfo{author}{Ghassemi, M.}
\newblock \bibinfo{title}{Chexclusion: Fairness gaps in deep chest x-ray
  classifiers} (\bibinfo{year}{2020}).
\newblock \eprint{2003.00827}.

\bibitem{selvaraju2017grad}
\bibinfo{author}{Selvaraju, R.~R.} \emph{et~al.}
\newblock \bibinfo{title}{Grad-cam: Visual explanations from deep networks via
  gradient-based localization}.
\newblock In \emph{\bibinfo{booktitle}{Proceedings of the IEEE international
  conference on computer vision}}, \bibinfo{pages}{618--626}
  (\bibinfo{year}{2017}).

\end{thebibliography}

% \noindent LaTeX formats citations and references automatically using the bibliography records in your .bib file, which you can edit via the project menu. Use the cite command for an inline citation, e.g.  \cite{Hao:gidmaps:2014}.

% For data citations of datasets uploaded to e.g. \emph{figshare}, please use the \verb|howpublished| option in the bib entry to specify the platform and the link, as in the \verb|Hao:gidmaps:2014| example in the sample bibliography file.

\iffalse
\section*{Acknowledgements (not compulsory)}

Acknowledgements should be brief, and should not include thanks to anonymous referees and editors, or effusive comments. Grant or contribution numbers may be acknowledged.
\fi

\section*{Author contributions statement}
Si.T. and So.T. conducted the experiments, So.T and N.F. wrote the manuscript, Si.T. provided the figures and plots, M.S. and M.H.R. provided critical feedback and suggestions. All authors reviewed the manuscript.

\section*{Additional information}

The authors declare no competing interests.

\noindent \textbf{Correspondence} and requests for materials should be addressed to M.H.R.

\iffalse

To include, in this order: \textbf{Accession codes} (where applicable); \textbf{Competing interests} (mandatory statement). 

The corresponding author is responsible for submitting a \href{http://www.nature.com/srep/policies/index.html#competing}{competing interests statement} on behalf of all authors of the paper. This statement must be included in the submitted article file.

\begin{figure}[ht]
\centering
\includegraphics[width=\linewidth]{stream}
\caption{Legend (350 words max). Example legend text.}
\label{fig:stream}
\end{figure}

\begin{table}[ht]
\centering
\begin{tabular}{|l|l|l|}
\hline
Condition & n & p \\
\hline
A & 5 & 0.1 \\
\hline
B & 10 & 0.01 \\
\hline
\end{tabular}
\caption{\label{tab:example}Legend (350 words max). Example legend text.}
\end{table}

Figures and tables can be referenced in LaTeX using the ref command, e.g. Figure \ref{fig:stream} and Table \ref{tab:example}.

\fi

\end{document}